\documentclass{article}

\usepackage{arxiv}

\usepackage[utf8]{inputenc} 
\usepackage[T1]{fontenc}    
\usepackage{hyperref}       
\usepackage{url}            
\usepackage{booktabs}       
\usepackage{amsfonts}       
\usepackage{nicefrac}       
\usepackage{microtype}      
\usepackage{lipsum}		

\usepackage{graphicx}
\usepackage[numbers]{natbib} 
\usepackage{doi}
\usepackage{authblk}
\usepackage{subcaption}

\title{Integrating Single-Cell Foundation Models with Graph Neural Networks for Drug Response Prediction}

\author[1,2]{Till Rössner}
\author[1,2]{Ziteng Li}
\author[1,2]{Jonas Balke}
\author[2]{Nikoo Salehfard}
\author[2]{Tom Seifert}
\author[1,2]{Ming Tang\textsuperscript{*}}

\affil[1]{L3S Research Center, Germany}
\affil[2]{Leibniz University Hannover, Germany}

\affil[ ]{\texttt{\{till.roessner, jonas.balke, nikoo.salehfard, tom.seifert\}@stud.uni-hannover.de}}
\affil[ ]{\texttt{\{ziteng.li, tang\}@l3s.de}}

\date{}

\renewcommand{\headeright}{}
\renewcommand{\undertitle}{}

\begin{document}

\maketitle

\footnotetext[1]{Corresponding author: \texttt{tang@l3s.de}}

\begin{abstract} 
AI-driven drug response prediction holds great promise for advancing personalized cancer treatment. However, the inherent heterogenity of cancer and high cost of data generation make accurate prediction challenging. In this study, we investigate whether incorporating the pretrained foundation model scGPT can enhance the performance of existing drug response prediction frameworks. Our approach builds on the DeepCDR framework, which encodes drug representations from graph structures and cell representations from multi-omics profiles. We adapt this framework by leveraging scGPT to generate enriched cell representations using its pretrained knowledge to compensate for limited amount of data. We evaluate our modified framework using IC$_{50}$ values on Pearson correlation coefficient (PCC) and a leave-one-drug out validation strategy, comparing it  against the original DeepCDR framework and a prior scFoundation-based approach. scGPT not only outperforms previous approaches but also exhibits greater training stability, highlighting the value of leveraging scGPT-derived knowledge in this domain.

\end{abstract}

\keywords{Drug Response Prediction \and Single-Cell Foundation Model \and Graph Neural Networks }

\section{Introduction}
Predicting Cancer Drug Response (CDR) is a critical challenge in cancer treatment due to the high heterogeneity of tumors and the vaiability of drug efficacy. These factors make accurate CDR prediction a complex and demanding task. Given its importance for understanding tumor behavior and guiding anticancer drug development, CDR predcition has become a major focus of reserach in recent years.

Prior works include network-driven methods, which build a similarity-based model and assign the sensitivity profile of an existing drug to a new drug if the structures of these two drugs are similar. Alternatively, machine learning approaches directly leverage large-scale drug and cancer cell line profiles to model drug response. However, both strategies have notable limitations. To address these, DeepCDR was introduced as a hybrid graph convolutional network that integrates multimodal data. As a deep learning–based framework, DeepCDR achieved state-of-the-art performance, surpassing earlier methods in accurately predicting cancer drug responses.\cite{liu2020deepcdr}

In a subsequent study, scFoundation - a transformer-based foundation model pretrained on large-scale single-cell data - was integrated into DeepCDR. By capturing complex contextual relationships among genes across diverse cell types, scFoundation enhanced CDR prediction through the provision of rich cell embeddings derived from single-cell gene expression data\cite{hao2024large}. This approach inspired us to explore the integration of another foundation model scGPT, which was pretrained using an alternative strategy and thus encodes cellular information in a distinct manner than scFoundation.

 scGPT is a generative pretrained transformer trained on 33 million single cells. It was primarily developed for tasks like cell type annotation, gene network inference, multi-batch and multi-omic integration \cite{cui2024scgpt}. However, its potential in CDR prediction has not yet been explored. In this study, we investigate whether the embeddings generated by scGPT can further improve the performance of DeepCDR in predicting drug responses.

\section{Related work}  
\subsection {Cancer Drug Response}  
Cancer Drug Response (CDR) is defined as the response of cancer cells or patients to a particular drug, which can be characterized as either sensitive or resistant. This response is typically quantified by the half-maximal inhibitory concentration IC$_{50}$, which is defined as the concentration of a drug required to inhibit a biological or biochemical function by 50\%. The IC$_{50}$ value is crucial for assessing the potency and efficacy of a drug in clinical settings, guiding therapeutic decisions and drug development strategies. Accurate prediction of CDR is essential in drug development, particularly in the field of personalized medicine, where treatments are tailored to an individual's genetic profile. The employment of AI methods for predicting IC$_{50}$ values has the potential to reduce the reliance on costly and time-consuming wet-lab experimental testing, thereby accelerating the identification of effective cancer therapies. 

\subsection{DeepCDR} DeepCDR is a cancer drug response prediction model designed to estimate the half-maximal inhibitory concentration (IC$_{50}$) of a drug for a given cancer cell line in regression tasks, or to classify the drug as sensitive or resistant in classification tasks. The model incorporates a graph convolutional network (GCN) to represent drug structures based on their chemical graphs, along with multiple subnetworks that extract features from multi-omics data—including genomics, transcriptomics, and epigenomics. \cite{liu2020deepcdr}




\subsection{Single-Cell Foundation Models}
scFoundation is a single-cell foundation model pretrained on over 50 million single-cell RNAseq data. It employs an asymmetric transformer architecture leveraging the sparsity of gene expression data to enhance its training efficiency and scalability. When integrated into the DeepCDR framework to generate cell embedding from gene expression data, scFoundation outperforms the original DeepCDR approach. \cite{hao2024large}

scGPT is another single-cell foundation model pretrained on over 33 million cells. Inspired by the success of self-supervised pretraining in natural language generation (NLG) using self-attention transformer, scGPT modifies the transformer architecture to learn cell and gene representations based on non-sequential omics data. By achieving state-of-the-art performance in multiple downstream task, it demonstrates its ability for transfer learning and capturing complex interactions across genes. \cite{cui2024scgpt}

\section{Methodology}

\subsection{Model Adaptation}

\begin{figure*} 
\centering \includegraphics[width=\textwidth]{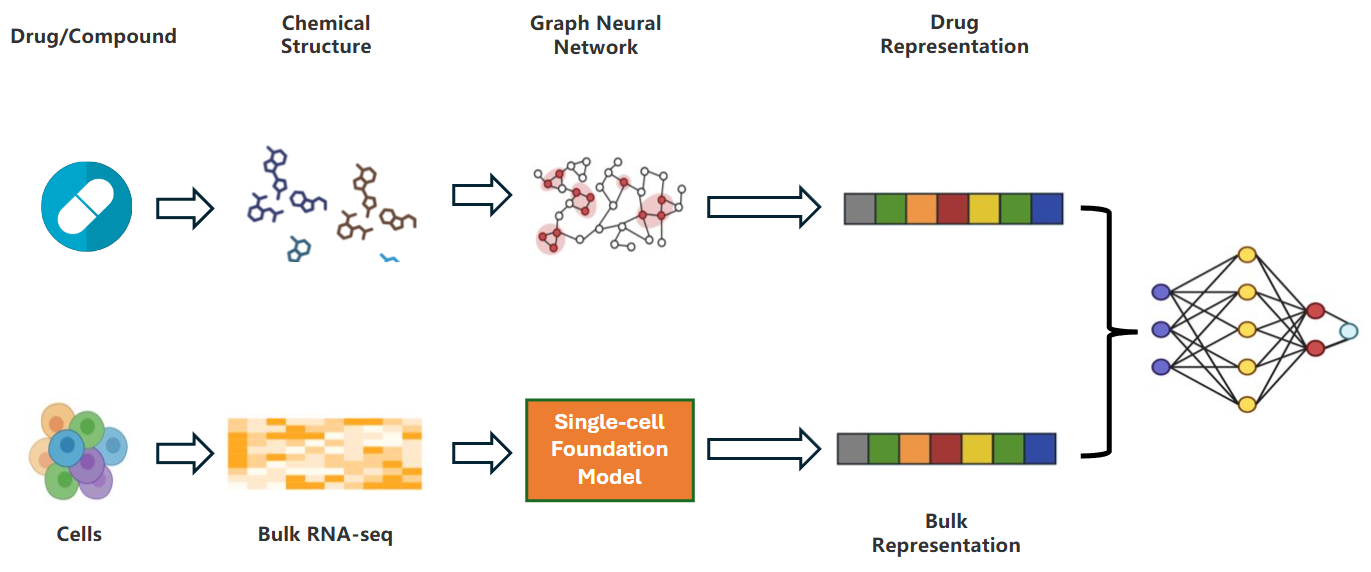} 
\caption{Overview of our model framework. Drug structures are encoded as molecular graphs and processed by a graph neural network. Cell bulk expression data are embedded using a foundation model. The resulting drug and cell representations are concatenated and used to predict drug response (IC$_{50}$) via a Neural Net.
To mitigate overfitting, we incorporate Dropout and Batch normalization. For classification sigmoid function was applied after the final layer.} 
\label{fig:DeepCDR} \end{figure*} 

As demonstrated in Figure \ref{fig:DeepCDR}, our model integrates the chemical structure of cancer drugs  with bulk RNA-sequencing data from cancer cell lines to predict the IC$_{50}$ value for a given drug - cell line pair. Each drug is represented by its molecular graph, capturing its chemical structure, while each cancer cell line is represented by its bulk gene expression pattern. Drug graphs are processed using a Graph Neural Network (GNN), an architecture designed to extract both local and global structural patterns from graph-structured data—features that are essential for modeling drug activity. A max pooling operation is then applied to summarize the most salient features from each molecular graph. On the cancer cell line side, gene expression data is embedded using the pretrained scGPT foundation model, which encodes rich representations of cellular states. These drug and cell line embeddings are concatenated, and passed through a neural net.
to predict IC$_{50}$ values - an indicator for drug sensitivity. Unlike the original DeepCDR framework, our approach omits methylation and mutation data, focusing exclusively on gene expression pattern to represent cell lines \cite{liu2020deepcdr}. Our implementation is based on the publicly available codebases of scFoundation and DeepCDR models, with necessary adaptations \cite{hao2024large} \cite{liu2020deepcdr}.

\subsection{Data Sources}
The data used in this study was sourced from the scFoundation work \cite{scfoundationgithub2024} \cite{hao2024large}, which compiled datasets from the Cancer Cell Line Encyclopedia (CCLE) \cite{Barretina2012CCLE} and the Genomics of Drug Sensitivity in Cancer (GDSC) \cite{Iorio2016GDSC}.

The CCLE dataset contains bulk RNA expression values of 697 genes for 561 cancer cell lines \cite{Barretina2012CCLE}. The GDSC dataset contains the ground truth values for model training, offering IC$_{50}$ values for specific drug–cell line pairs corresponding to those in the CCLE dataset \cite{Iorio2016GDSC}.Each drug is represented as a molecular graph composed of three components: a feature matrix (75-dimensional) encoding the attributes of each atom in the molecule, an adjacency list specifying the bonds between atoms based on their indices, and a degree list indicating the number of neighbors for each atom \cite{liu2020deepcdr}.

To ensure compatibility with scGPT and scFoundation, which expect a broader set of input genes, we used a gene list provided by scFoundation \cite{scfoundationgithub2024} and applied zero-padding for genes absent from the CCLE gene expression dataset but present in the expected input set.

\subsection{Creation of the embeddings}
Our embeddings were created using a pretrained scGPT model. In detail, the scGPT-human checkpoint from the scGPT zero-shot tutorial \cite{scgptgithub2025} was used. To maintain consistency in environmental factors during embedding generation, the gene expression data underwent preprocessing steps aligned with those used in scFoundation. This included zero-padding for genes present in the expected input gene list but absent from the expression dataset, as well as normalization using counts per million (CPM) followed by a log1p transformation to stabilize variance and reduce the influence of extreme values. The embeddings generated by scGPT had a dimensionality of 512, whereas those produced by scFoundation had a dimensionality of 768. This difference in embedding size was acknowledged and accounted for in the comparative analysis of the performance of the two models. In general, higher-dimensional embeddings, can potentially capture more features of the input data, at the cost of increased computational requirements.


\begin{figure*}
    \centering
    \includegraphics[width=0.9\linewidth]{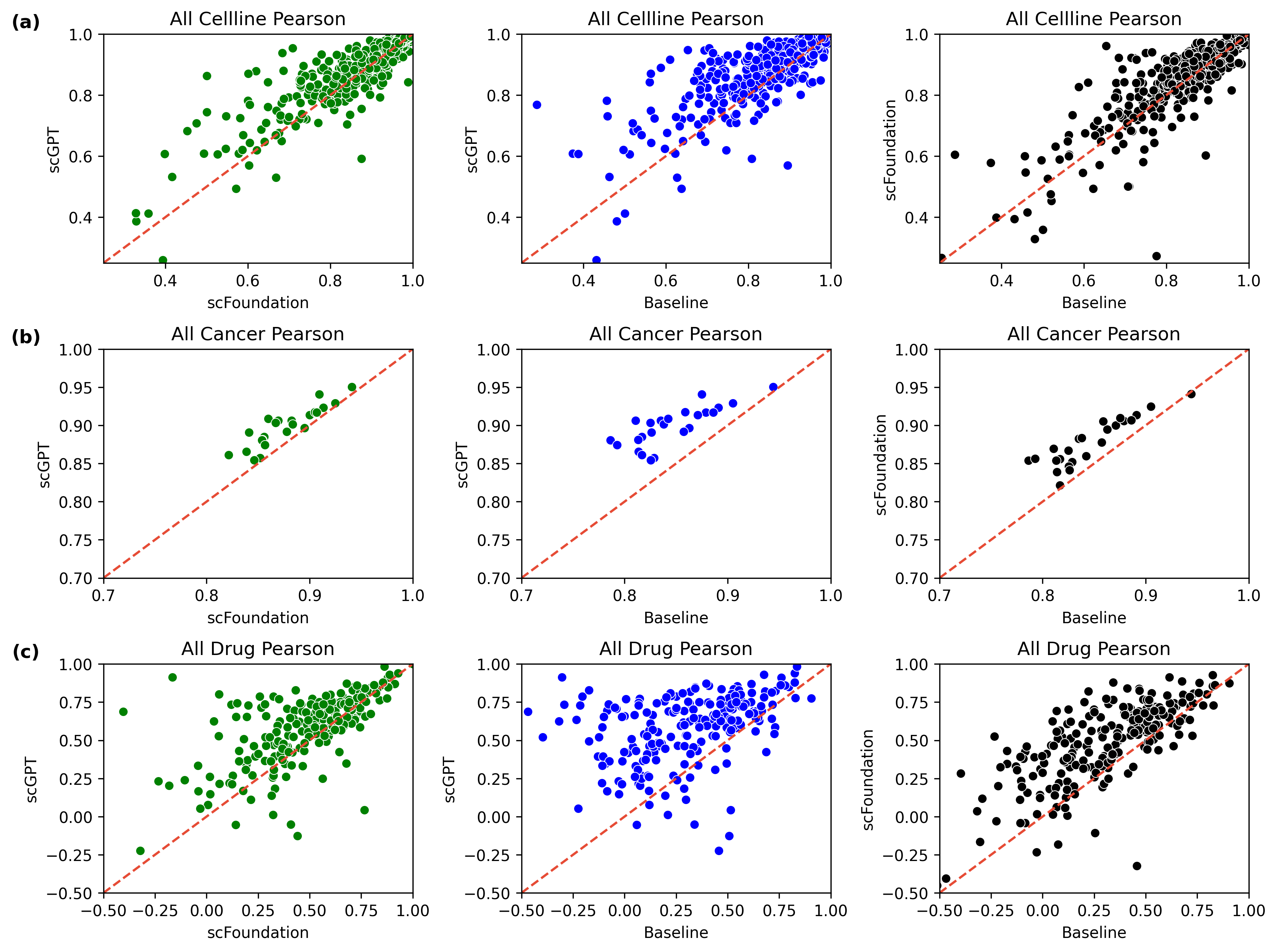}
    \caption{Comparison of scGPT, scFoundation and Baseline results: PCC values between predicted and actual responses for (a) all cell lines, (b) all cancer types, (c) and all drugs in the test set were illustrated. Each dot representing a distinct cell line, cancer type, or drug. The red dashed identity line (y = x) serves as a reference for assessing relative model performance.}
    \label{fig:Results1}
\end{figure*}

\subsection{Evaluation}
We adopted evaluation metrics consistent with those used in scFoundation \cite{hao2024large}. In all experiments, 95\% of the data was used for training, with the remaining 5\% reserved for testing. Due to memory constraints, we were not able to include the full dataset into the training process and limited the number of training instances to 90000. This subset was selected by slicing rather than random sampling, which may have introduced bias and affected the evaluation results.

\textbf{Baseline:}
To evaluate the effectiveness of our proposed approach, we compare it against two baselines: the original DeepCDR framework and an adapted scFoundation-based model. For consistency, both baselines were modified to exclude methylation and mutation data, relying solely on gene expression inputs.

\textbf{Pearson Correlation Coefficient (PCC):}
The model's performance was primarily evaluated using the PCC, which measures the linear relationship between the predicted drug responses and the observed ones. The evaluation was conducted at several filter levels (cell line, cancer type, drug), meaning all the predicted values for a given category were checked for their correlation to the corresponding true labels. 

\textbf{Leave-one-drug-out:}
Additionally, we conducted a series of leave-one-drug-out evaluations. In each run, the model was trained on nearly the entire dataset, with all samples associated with one held-out drug moved to the test set. This setup simulates a realistic scenario in which the model encounters a completely unseen drug. While the dataset includes 223 drugs in total, computational limitations restricted us to evaluating 20 randomly selected drugs for this analysis.

\textbf{Training stability:}
During training, we observed notable differences in training stability across the different model setups. To systematically assess this, we analyzed validation scores recorded at each training epoch for all three configurations. This allowed us to evaluate and compare the consistency and stability of the training process for each approach.



\section{Results and Discussion}  

\begin{figure}[h]
  \centering
  \begin{subfigure}[b]{0.48\textwidth}
    \centering
    \includegraphics[width=\linewidth]{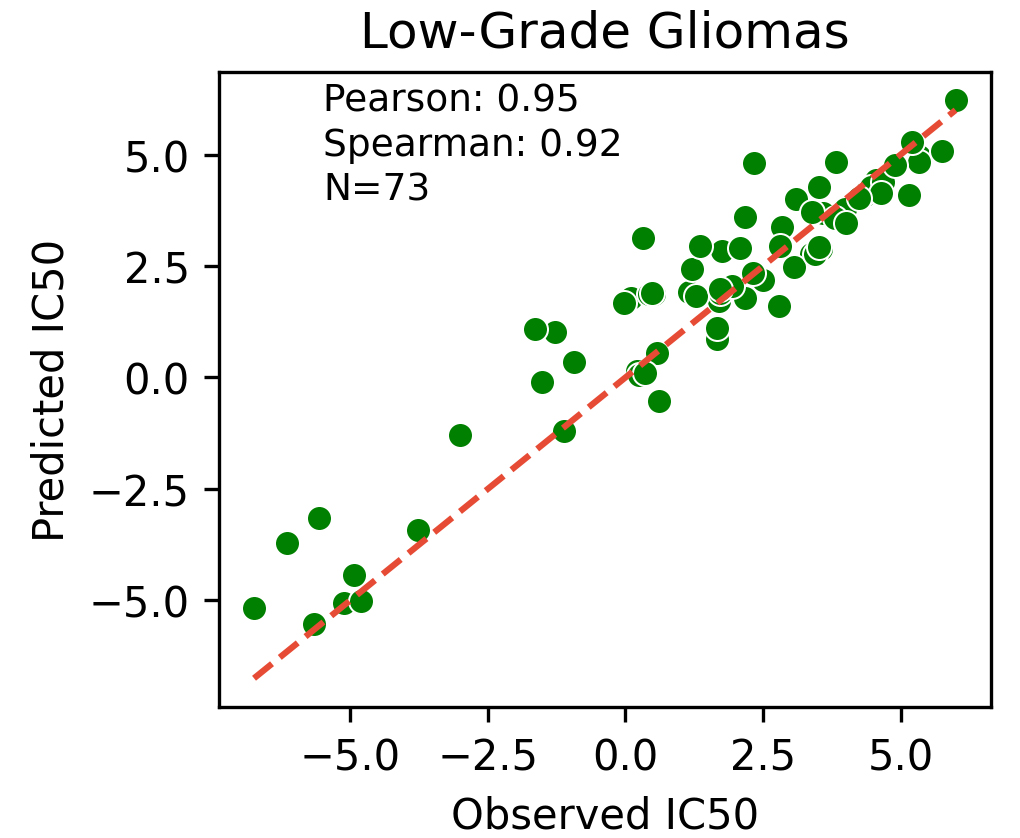}
    \caption{Low-Grade Gliomas}
    \label{fig:gliomas}
  \end{subfigure}
    \hfill
  \begin{subfigure}[b]{0.48\textwidth}
    \centering
    \includegraphics[width=\linewidth]{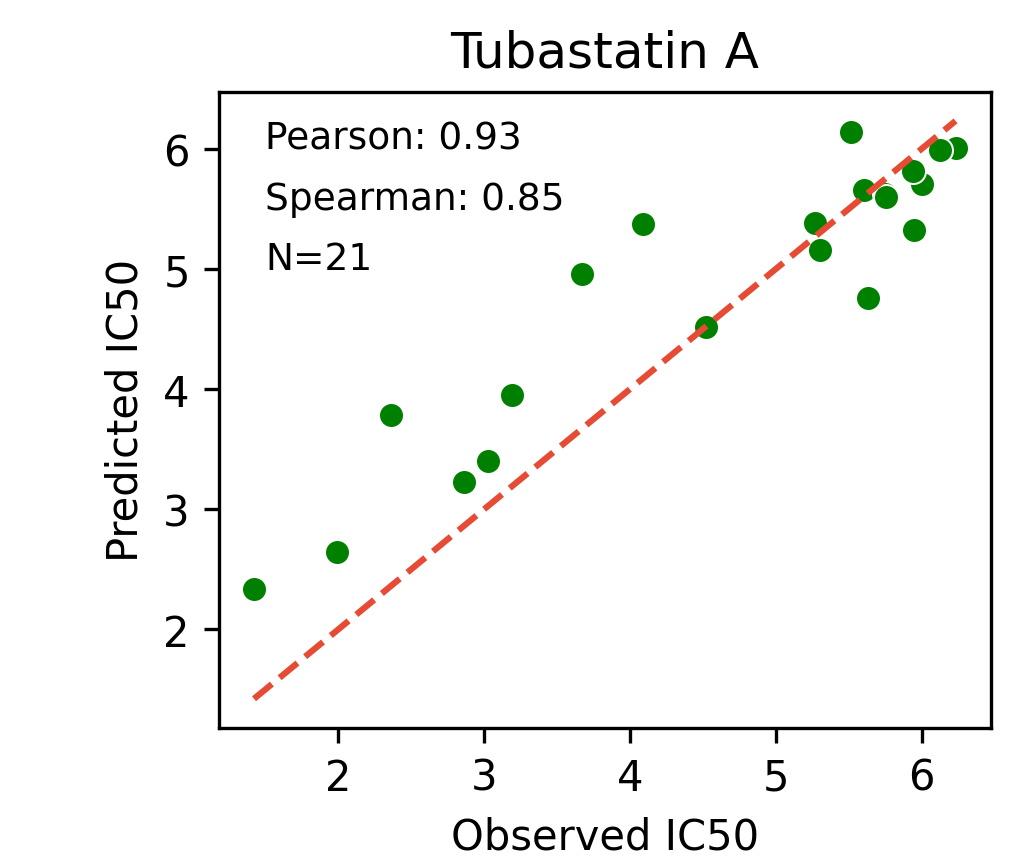}
    \caption{Tubastatin A}
    \label{fig:tubastatin}
  \end{subfigure}
  \caption{Relationship between predicted and observed IC$_{50}$ values for the best prediction case among cancer types, Low-Grade Gliomas (a), and drug types, Tubastatin A (b). Each dot represents a combination of a drug and a cell line, with the dashed red line indicating a perfect correlation between predicted and observed IC$_{50}$.}
  \label{fig:ic50_comparison}
\end{figure}
The proposed scGPT-based approach exhibited superior performance in predicting CDR in comparison to both the scFoundation-based model and the original DeepCDR method. As shown in Figure \ref{fig:Results1}, Pearson correlation coefficient (PCC) comparisons across all three evaluation settings — cell line-based, cancer type-specific, and drug-specific —reveal that the scGPT-based model outperforms the DeepCDR baseline by a substantial margin and exceeds the performance of scFoundation by a small but consistent degree. These results highlight the effectiveness of scGPT embeddings in capturing relevant cellular features, thereby enhancing the accuracy of IC$_{50}$ predictions. To more directly illustrate the prediction accuracy of IC$_{50}$, Figure \ref{fig:ic50_comparison} presents the correlation between predicted and observed IC$_{50}$ values, using low-grade glioma and the drug Tubastatin A as representative examples.


\begin{figure}[h]
    \centering
    \includegraphics[width=0.48\linewidth]{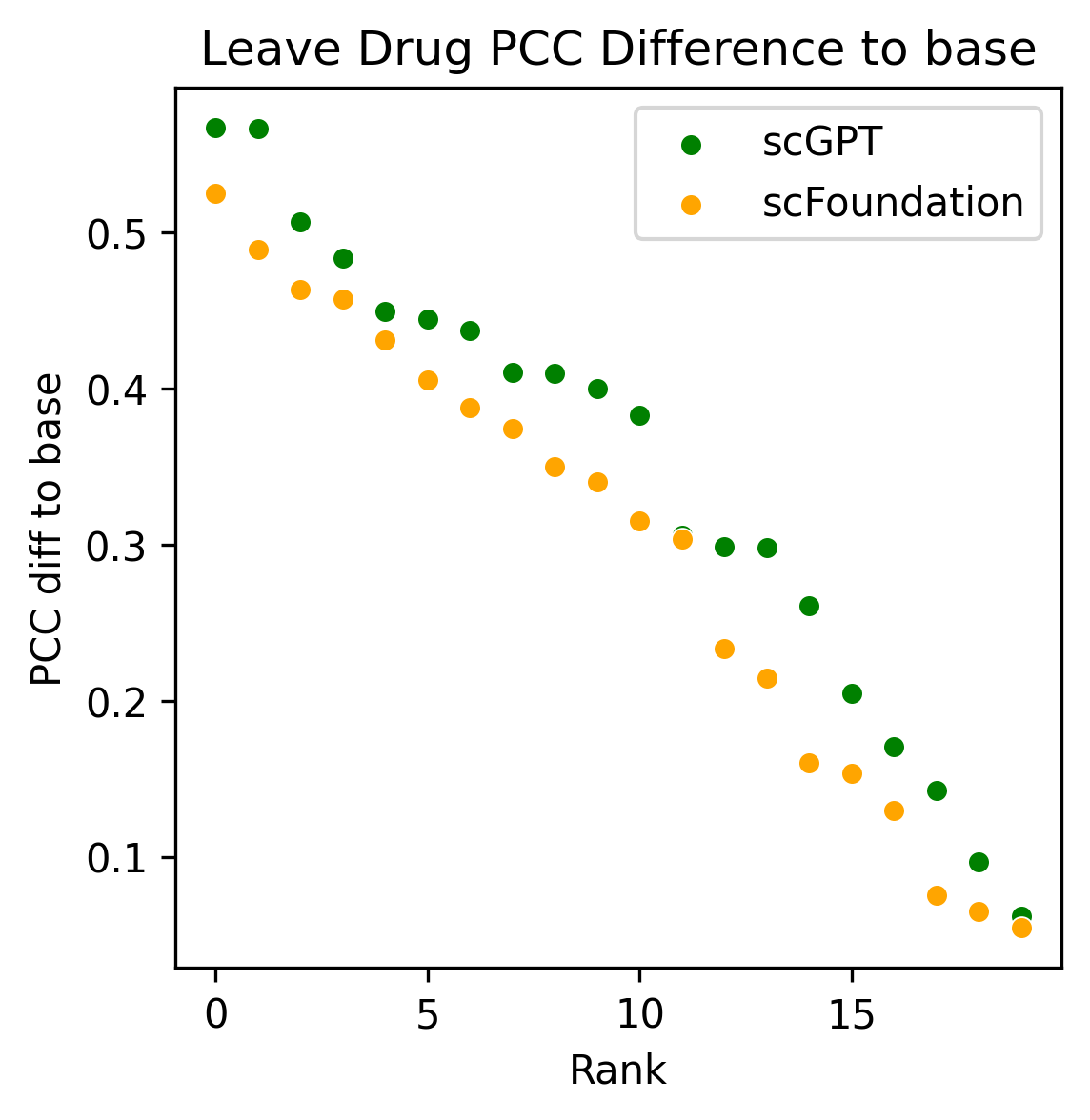}
    \caption{Comparison of models' performance in the leave-one-drug-out test: PCC value difference between two models—scGPT and scFoundation—compared to the baseline in a leave-one-drug-out test. Each dot represents one of the 20 randomly selected drugs from the dataset, with the y axis indicating the gained PCC values and the x axis representing the rank of drugs based on their improvement.}
    \label{fig:results3}
\end{figure}

In addition, an evaluation of the approach was conducted using the leave-one-drug-out test. In this test, we removed one drug entirely from the training dataset, train the model without it, and then evaluated the model on the excluded drug to assess its ability to predict results for unseen drugs. In the blind drug test, our scGPT-based approach outperformed both: the scFoundation and baseline model. The figure \ref{fig:results3} demonstrates that scGPT consistently achieves higher PCC gains across various drugs compared to the scFoundation method, underscoring its enhanced predictive capability. Notably, the original DeepCDR model struggled to generalize to unseen drugs, highlighting the importance of foundation models, which leverage large-scale single-cell data to learn robust gene representations \cite{hao2024large} \cite{cui2024scgpt}.
The limitation of the training data to 90000 instances did not seem to have too big of an impact on the training performance as the general range of the results is still similar to the results from the experiments of the scFoundation DeepCDR evaluation \cite{hao2024large}.

\begin{figure}
    \centering
    \includegraphics[width=0.8\linewidth]{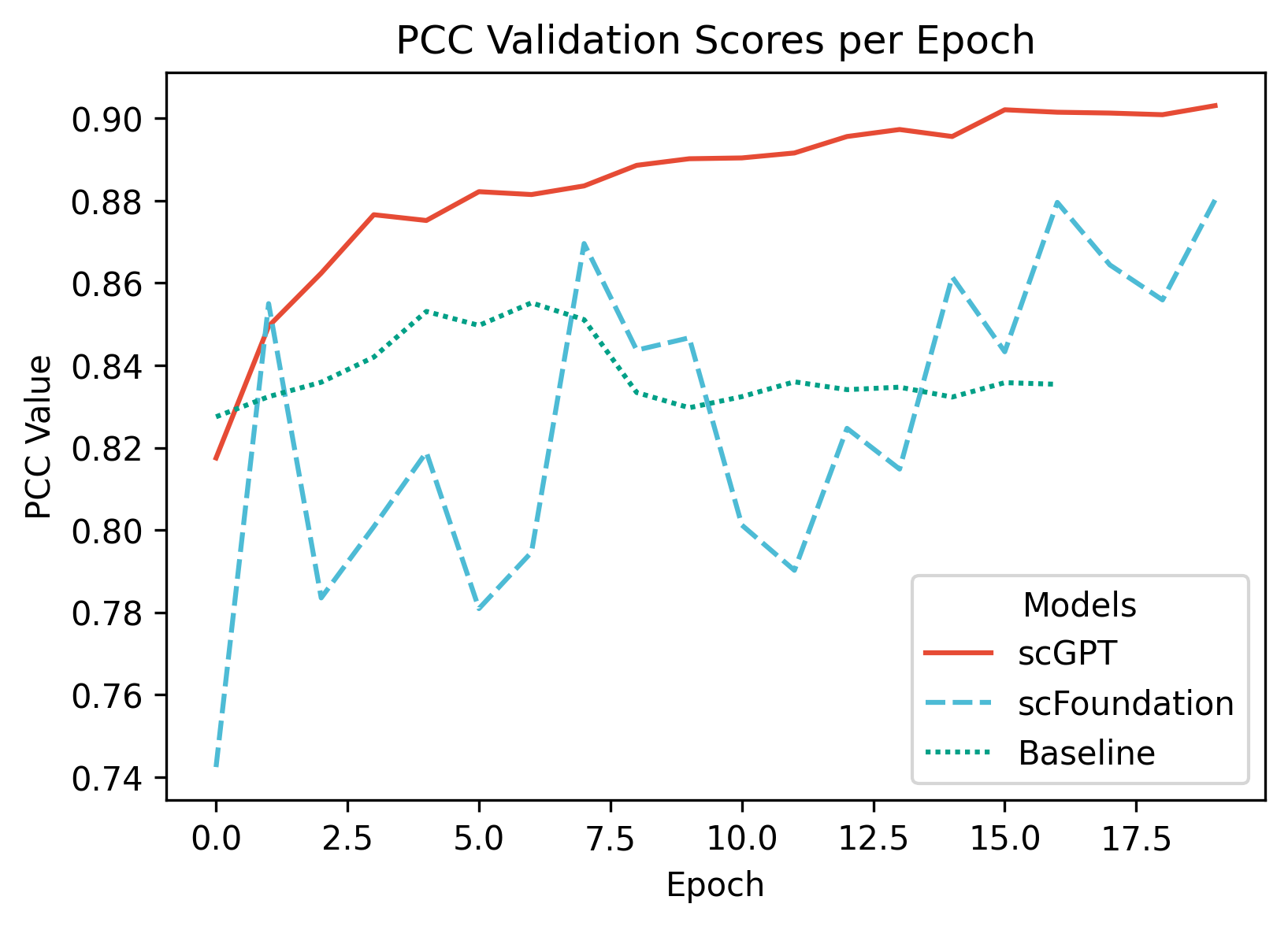}
    \caption{Comparison of PCC values per Epoch for scGPT-based model, scFoundation-based model, and the baseline over 20 training epochs (baseline was stopping early after 17 epochs).}
    \label{fig:Results4}
\end{figure}

Another evaluation we conducted was on the stability of the models during the training process. Figure \ref{fig:Results4} illustrates that scGPT-based model achieves steady and consistent performance improvement, starting at a PCC of approximately 0.80 and surpassing 0.90 over the course of training. This indicates superior training stability compared to the other models. In contrast, the scFoundation model shows substantial fluctuations, suggesting lower stability. The baseline model remains relatively stable throughout the training but consistently underperforms compared to scGPT. Overall, scGPT not only achieves the highest final PCC value but also maintains a smooth and stable learning trajectory.

\section{Conclusion}  

In this study, we proposed a novel approach that integrates scGPT embeddings into the DeepCDR framework for predicting cancer drug responses. While we initially expected scGPT to perform comparably to scFoundation—especially given its lower embedding dimensionality—our experimental results consistently showed that the scGPT-based model outperforms both scFoundation and the original DeepCDR across multiple evaluation settings.

Pearson correlation analyses across cell lines, cancer types, and drugs demonstrated that the scGPT model more accurately predicts IC$_{50}$ values, indicating a stronger ability to capture the complex relationships between drug structures and gene expression profiles. In leave-one-drug-out tests, scGPT also showed superior generalization to previously unseen drugs—an essential capability for real-world applications like computational drug candidate evaluation, which the original DeepCDR model lacks. Moreover, scGPT exhibited greater training stability, with a smooth and consistent improvement in validation performance throughout training, outperforming both comparison models in consistency and final accuracy.

These findings highlight the potential of scGPT-derived bulk RNAseq embeddings as a powerful alternative to traditional approaches, advancing the field of cancer drug discovery and personalized medicine.

\section{Outlook}
In this study, we focused on enhancing cell representation by leveraging the transfer learning ability from foundation models, specifically scGPT. We did not explore adaptations to the drug representation, which presents a promising direction for future research. In particular, incorporating foundation model tailored to drugs may yield further performance gains. Recent advancements in molecular embeddings fall into two primary categories: (1) NLP-based methods that encode molecules as SMILES strings, and (2) graph-based approaches that represent molecular structures directly as graphs. State-of-the-art models like ChemBERTa-2 \cite{ahmad2022chemberta2}, SPMM \cite{chang2024spmm}, and MOLFormer \cite{ross2022molformer} apply transformer architectures to SMILES strings. Conversely, models such as MolE \cite{mendez2024mole} and GROVER \cite{rong2020grover} operate on molecular graphs and achieve leading results across various downstream tasks. These models, trained on millions of molecules, offer rich pretrained representations that could be integrated into DeepCDR to further enhance drug response predictions.

Additionally, reincorporating methylation and mutation data into the adapted DeepCDR framework represents another valuable research direction. Prior work by Liu et al. \cite{liu2020deepcdr} demonstrated the benefit of multi-omics integration for improving cancer drug response prediction.



\bibliographystyle{unsrtnat}
\bibliography{reference}  






\end{document}